\documentclass[runningheads]{llncs}
\usepackage[T1]{fontenc}
\usepackage{graphicx,verbatim}
\usepackage{cite}
\usepackage{url}
\usepackage{amsmath,amsfonts}
\usepackage{booktabs}
\usepackage{multirow}
\usepackage[
  breaklinks=true,
  bookmarks=false,
  colorlinks=true,
  linkcolor=blue,
  urlcolor=blue,
  citecolor=blue,
]{hyperref}
\usepackage[capitalize]{cleveref}
\crefname{section}{Sec.}{Sec.}
\crefname{table}{Tab.}{Tab.}
\crefname{figure}{Fig.}{Fig.}

\usepackage{tikz}
\usepackage{xcolor}
\usepackage{xspace}
\newif\ifanonymized
\anonymizedfalse     

\newcommand{\anonymize}[2]{\ifanonymized#2\else#1\fi}

\newcommand{\IKEM}{\anonymize{IKEM}{AH-A}\xspace}
\newcommand{\FTN}{\anonymize{TUH}{AH-B}\xspace}
\newcommand{\KNL}{\anonymize{KNL}{AH-C}\xspace}
\newcommand{\IKEMfull}{\anonymize{Institute of Clinical and Experimental Medicine, Prague}{Anonymised Hospital~A}\xspace}
\newcommand{\FTNfull}{\anonymize{Thomayer University Hospital, Prague}{Anonymised Hospital~B}\xspace}
\newcommand{\KNLfull}{\anonymize{Regional Hospital Liberec}{Anonymised Hospital C}\xspace}

\definecolor{minblue}{RGB}{247, 251, 255}
\definecolor{maxblue}{RGB}{8, 48, 107}

\newcommand{\confusionmatrix}[4]{
    \begin{scope}[shift={(#1,#2)}]
        \node[font=\normalsize] at (2, 1) {#3};
        
        \foreach \x in {0,1,2,3,4} {
            \node[font=\scriptsize\sffamily] at (\x, -4.7) {\x};
            \node[font=\scriptsize\sffamily] at (-0.7, -\x) {\x};
        }

        \foreach \x/\y/\val in {#4} {
            \pgfmathsetmacro{\pct}{\val * 100}
            
            \pgfmathparse{\val > 0.49 ? "white" : "black"}
            \edef\txtcolor{\pgfmathresult}
            
            \fill[maxblue!\pct!minblue] (\x-0.5, -\y+0.5) rectangle (\x+0.5, -\y-0.5);
            
            \node[font=\scriptsize\sffamily, text=\txtcolor] at (\x, -\y) {\val};
        }
    \end{scope}
}
\usepackage{orcidlink}
\renewcommand{\orcidID}[1]{\orcidlink{#1}}

\begin{document}
\title{Weakly Supervised Multicenter Nancy Index Scoring in Ulcerative Colitis Using Foundation Models}
\titlerunning{Weakly Supervised Multicenter Nancy Scoring}

\ifanonymized
\author{Anonymized Authors}  
\authorrunning{Anonymized Author et al.}
\institute{Anonymized Affiliations \\
    \email{email@anonymized.com}}
\else
\author{Adam Kukučka\inst{1}\orcidID{0009-0002-3341-9441} \and
Ondřej Fabián\inst{2}\orcidID{0000-0002-0393-2415} \and
Vít Musil\inst{1}\orcidID{0000-0001-6083-227X} \and
Tomáš Brázdil\inst{1}\orcidID{0000-0002-4547-3261}}
\authorrunning{A.\ Kukučka et al.}
\institute{Masaryk University, Faculty of Informatics, Brno, Czech Republic
\and
Institute of Clinical and Experimental Medicine, Prague, Czech Republic}
\fi

\maketitle
\begin{abstract}
Histologic assessment of ulcerative colitis (UC) activity is an important endpoint in clinical trials and routine care, but manual grading with indices such as the Nancy histological index (NHI) is time-consuming and prone to observer variability.
While computational pathology methods can automate scoring, many approaches depend on dense region-level annotations, which are costly to obtain, particularly in heterogeneous, multicenter cohorts.
We propose a weakly supervised multiple instance learning (MIL) approach for whole-slide images that learns from case- and slide-level NHI labels, leveraging foundation models.
Our method targets clinically relevant endpoints, including neutrophilic activity and derived Nancy-low/high groupings, enabling full five-grade NHI prediction.
On a multicenter dataset of H\&E-stained colon biopsies from three hospitals (2019--2025), we evaluate multiple foundation model encoders and aggregation strategies.
We find that foundation model choice and resolution substantially affect performance, with Virchow2 providing the most consistent gains, and that a simple ensembling rule improves five-grade NHI prediction compared to a hierarchical gating baseline.
Overall, our results demonstrate that weakly supervised MIL with modern foundation-model representations can provide robust, interpretable UC histology activity assessment in realistic multicenter settings.
\end{abstract}

\section{Introduction}

Inflammatory bowel disease (IBD) is a chronic systemic inflammatory disorder primarily affecting the gastrointestinal tract and comprises two main subtypes: ulcerative colitis (UC) and Crohn’s disease.
UC is characterised by continuous chronic inflammation limited to the colonic mucosa.
Histopathological assessment remains central to both diagnosis and evaluation of inflammatory activity.

In clinical trials and increasingly in routine care, microscopical disease activity is graded using standardized histologic indices, e.g., the Nancy histological index (NHI)~\cite{Marchal-Bressenot43}, which operationalize the presence and severity of key microscopic findings, such as infiltration.
However, manual grading is time-consuming and subject to inter- and intra-observer variability, especially when applied at scale in multicenter studies~\cite{Geboes404}.

Computational pathology methods have shown promise for automating histologic scoring~\cite{10.1093/ecco-jcc/jjae198}, yet many existing approaches rely on dense, region-level annotations that are expensive to obtain and difficult to standardize across institutions.
In this work, we focus on a realistic setting in which only case-/slide-level labels are available, and the training data originate from multiple centers with heterogeneous staining, scanning protocols, and patient populations.

\paragraph{Problem Formulation.}
For a whole-slide image (WSI) $X$, let $y \in \{0,1,2,3,4\}$ denote its NHI grade.
Our primary objective is to learn a predictor $X\mapsto \hat y$.
We consider a weakly supervised setting in which only slide- or case-level labels $y$ are available, with no region- or cell-level annotations.
\medskip

\paragraph{Our Contributions.}
To address this weakly supervised, multicenter scenario, we develop a multiple instance learning (MIL) pipeline that operates on tiled whole-slide images, integrates quality control and tissue filtering, and leverages modern visual representations to improve robustness.
We evaluate our approach on a multicenter cohort and report performance for clinically relevant endpoints, including neutrophilic activity and derived NHI groupings.
In summary, our contributions are:

\begin{itemize}
    \item \textbf{Weak annotations.}\quad
    We tackle the Nancy histological index prediction task under realistic weak supervision, using only case- and slide-level labels rather than dense region-level annotations.
    \item \textbf{MIL architecture.}\quad
    We propose a weakly supervised MIL approach tailored to UC histology, combining foundation-model encoders with task-specific aggregation to predict clinically relevant endpoints, including neutrophilic activity and derived NHI low/high groupings
    \item \textbf{Multicenter evaluation.}\quad
    We evaluate our approach across three institutions with heterogeneous staining protocols.
\end{itemize}

\subsection{Related Work}

\paragraph{Histologic Indices for UC.}
Histologic disease activity in UC is commonly assessed with standardized grading systems such as the Geboes score, the NHI, and the Robarts histopathology index~\cite{Geboes404,Marchal-Bressenot43,Mosli50}, which are widely used clinical and research endpoints but remain time-consuming to apply at scale.

\paragraph{AI for UC/IBD Histology.}
AI-based assessment has progressed from proof-of-concept to multicenter validation, with a recent meta-analysis showing performance comparable to expert pathologists for histological remission assessment~\cite{10.1093/ecco-jcc/jjae198}.
Multiple studies report feasibility for UC histology assessment and clinical use~\cite{OHARA2025846,10.1093/ecco-jcc/jjab232.014,Iacucci2023,Furlanello2025}.
Several approaches explicitly model NHI targets or support NHI scoring in multicenter settings, including granular supervision and assistive tools, as well as feature-specific quantification (e.g., eosinophils) linked to clinicopathologic variables~\cite{Najdawi2023,10.1093/ibd/izae204,https://doi.org/10.1002/ueg2.12562,10.1093/ibd/izab122}.
Endoscopy-linked activity prediction has also been explored~\cite{Bossuyt1778}, and weakly supervised WSI approaches provide a path to leverage coarse labels in IBD datasets~\cite{mokhtari2023interpretablehistopathologybasedpredictiondisease}.

\paragraph{Weakly Supervised WSI Learning.}
WSI analysis is commonly formulated as MIL, where attention pooling is a standard aggregation mechanism and large-scale weakly supervised systems have achieved clinical-grade performance~\cite{ilse2018attentionbaseddeepmultipleinstance,Campanella2019}.
Extensions include transformer-based aggregators, dual-stream and contrastive objectives, pseudo-labeling/instance selection, and relational/multi-task designs~\cite{shao2021transmiltransformerbasedcorrelated,li2021dualstreammultipleinstancelearning,Zhou_2023,9695281,cancers14235778}.
Recent benchmarks and reviews summarize self-supervision gains and design choices across encoders, pretraining, and pooling, and highlight optimization details for weakly annotated WSI classifiers~\cite{mammadov2025selfsupervisionenhancesinstancebasedmultiple,ZHANG2025103027,SAEED2025109649}.

\paragraph{Foundation Models and Transferable Representations.}
Foundation models can reduce annotation needs and improve robustness across stain/scan variability; recent reviews summarize challenges and opportunities for clinical impact~\cite{bilal2025foundationmodelscomputationalpathology,khan2025comprehensivesurveyfoundationmodels}. Large-scale pathology pretraining now spans tile- and slide-level objectives, enabling transferable representations for downstream histology tasks~\cite{Xu2024,Chen2024,Vorontsov2024,zimmermann2024virchow2scalingselfsupervisedmixed,BILAL2026102680}.

\section{Material and Methods} 
\subsection{Data}
We retrospectively collected hematoxylin and eosin (H\&E) stained colon biopsies from routine operations between 2019 and 2025 from three institutions in the \anonymize{Czech Republic}{Anonymised country}, namely
\emph{\IKEMfull} (\IKEM);
\emph{\FTNfull} (\FTN);
and
\emph{\KNLfull} (\KNL).
All glass slides were digitized at 20\texttimes magnification (scanner: ZEISS Axioscan 7).
We included only slides from the colon and excluded ileal specimens.

Each case/slide was assigned a histologic activity grade using the NHI.
Due to differences in local data organization, label granularity differed across centers: \IKEM provides predominantly case-level labels, with most cases containing a single slide (on average 1.15 slides per case); \FTN provides exactly one slide per case (effectively slide-level labels); and \KNL contains multiple slides per case but provides slide-level labels.

Dataset splits were performed at the case level to prevent any patient/case from appearing in more than one split.
We separate training from two evaluation sets: a preliminary test set for model selection/ablation checks and a final test set for locked, unbiased reporting.
The total number of slides per institution was \IKEM 2017, \FTN 1339, and \KNL 176. \IKEM and \FTN were split in a 70:15:15 train:preliminary:test-final ratio, while \KNL used a 0:50:50 split between preliminary and final test sets.
Overall, the Nancy grade distribution is imbalanced (with more low-grade cases than high-grade), and class frequencies differ by hospital, motivating a multicenter evaluation.

\subsection{Pipeline}
\begin{figure}[t]
\centering
\includegraphics[width=\linewidth]{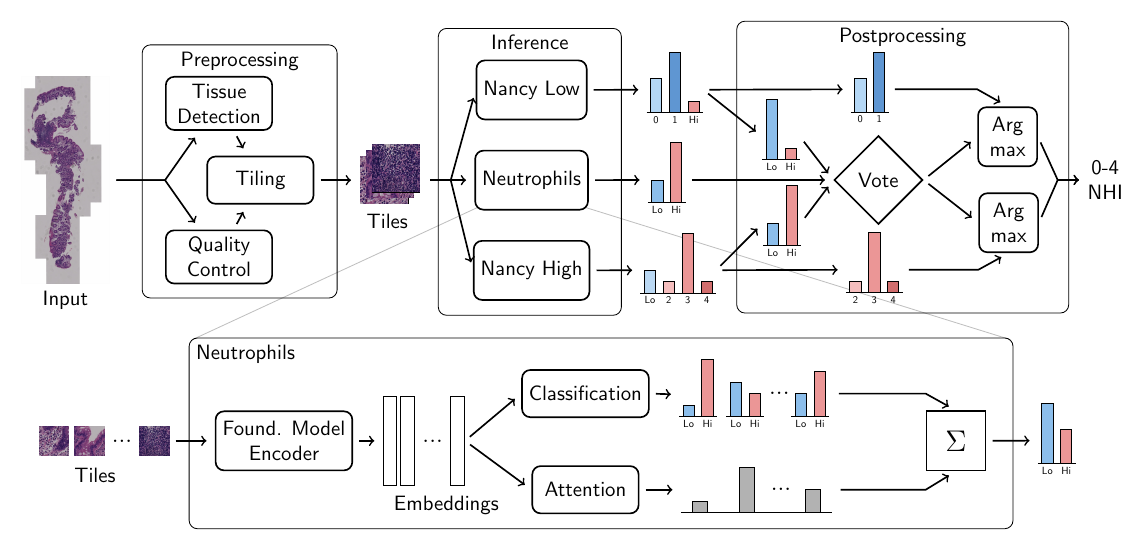}
\caption{Overview of the proposed pipeline, illustrating preprocessing (tissue detection, quality control, tiling), inference with a frozen foundation-model encoder and MIL aggregation, and post-processing to obtain the final NHI grade.}
\label{fig:pipeline}
\end{figure}

Figure~\ref{fig:pipeline} summarizes the full workflow.
In preprocessing, each WSI is converted into a bag of tiles.
For inference, the bag of tiles is processed by three models specialized to different NHI activity groups, each producing a probability distribution.
Finally, a post-processing step combines these three distributions into a single five-grade NHI prediction for the WSI.

\newcommand{\Lo}{\text{Lo}}
\newcommand{\Hi}{\text{Hi}}

\paragraph{Inference Models.}
The inference pipeline consists of three specialized, slide-level models that process a bag of $N$ tiles $x_i$ and output the distribution of the Nancy Histological Index (NHI).
To optimize accuracy across the scale, the NHI grades are partitioned into two groups: \Lo~(0–1) and \Hi~(2–4).
Each model serves a specific diagnostic role.
Namely, the first one predicts neutrophil activity, a clinically meaningful indicator of active inflammation, with output distribution over \{\Lo,\Hi\}.
The second model specializes in the Nancy-low regime and outputs distr.\ over \{0,1,\Hi\}.
In contrast, the third one focuses on Nancy-high grades (distr.\ over \{\Lo,2,3,4\}).

\paragraph{Tile Embeddings.}
Each tile is embedded by a frozen pretrained visual encoder~$f$ (a foundation model) into a $d$-dimensional representation $h_i = f(x_i) \in \mathbb{R}^d$.

\paragraph{Attention-based Aggregation.}
Each of the three inference models uses attention-based multiple instance learning (MIL), where tile embeddings are pooled into slide-level logits via task-specific attention weights (inspired by Ilse \emph{et al.}\cite{ilse2018attentionbaseddeepmultipleinstance}).
Specifically, the unnormalized attention score for tile $i$ is computed as
$u_i = w^{\top}\tanh\big(V h_i\big)$,
where $V\in\mathbb{R}^{m\times d}$ and $w\in\mathbb{R}^{m}$ are learnable parameters. 
Attention weights $(a_1,\ldots,a_N)=\operatorname{Softmax}(u_1,\ldots,u_N)$ are then obtained by a normalization over the bag.

\paragraph{Instance-first Classification (our variant).}
In a standard attention-MIL classifier, the slide embedding is first computed as $\sum_i a_i h_i$, and a classifier is applied on top.
Instead, we apply a task-specific classifier $c_\psi$ to each tile embedding first
to obtain tile-level logits $s_i = c_{\psi}\big(h_i\big)$.
We then obtain slide-level logits by attention-weighted pooling $s = \sum_{i} a_i\, s_i$.
The final slide-level prediction is obtained by applying the appropriate link function: a sigmoid for the binary neutrophil task and a softmax for the multi-class Nancy-low and Nancy-high tasks.

\paragraph{Post-processing with Ensembles.}
Our target label space is the full five-grade NHI, but we train three task-specific models: A binary neutrophil activity model distinguishing between \Lo{} and \Hi;
a Nancy-low model over \{0,1,\Hi\};
and a Nancy-high model over \{\Lo,2,3,4\}.

At inference time, we first determine the activity group \{\Lo,\Hi\} by majority voting across the three models.
We then delegate the final grade prediction to the appropriate specialist model.
The final grade is chosen as the $\arg\max$ over the specialist model's specific grade classes (i.e., the \Lo{} or \Hi{} class is ignored).

\paragraph{Preprocessing.}
We detect tissue regions (Otsu thresholding in HSV) with basic morphological cleanup, then apply quality control to remove low-quality tiles (e.g., blur/artifacts). Tiles are extracted at three resolutions:
(T0) level 0 (0.17\textmu m/px), 320\texttimes 320px, no overlap;
(T1) level 1 (0.52\textmu m/px), 224\texttimes 224 px, 112 px overlap;
(T2) level 2 (1.55\textmu m/px), 224\texttimes 224 px, 112 px overlap.
Implemented primarily in \anonymize{\texttt{ratiopath}\footnote{\url{https://github.com/RationAI/ratiopath}}}{Anonymised tool/repository}.

\paragraph{Training Details.}
Each slide corresponds to one bag and is supervised only by its slide-/case-level label (\IKEM slides inherit case labels); each bag contains all available tiles.
We used Adam (lr $10^{-5}$, batch size 8) for up to 300 epochs with early stopping (patience 20).
Losses were binary cross-entropy (neutrophil) and cross-entropy (Nancy-low/high) with class-balanced batches via over/undersampling.
The pretrained encoder was frozen; only attention $(V,w)$ and head $\psi$ were optimized, with no additional data augmentation.
We used five-fold cross-validation on the training set.

\section{Results and Discussion}

\subsection{Performance Evaluation}
\label{sec:results_by_center}

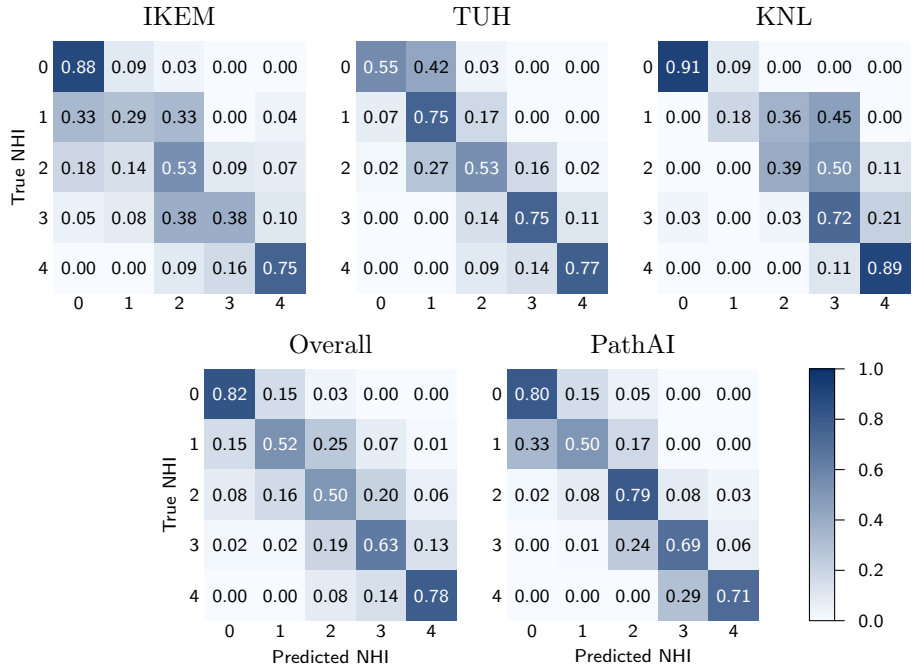
\begin{figure}[t]
\begin{tikzpicture}[scale=0.67]

    \confusionmatrix{0}{0}{\IKEM}{
        0/0/0.88, 1/0/0.09, 2/0/0.03, 3/0/0.00, 4/0/0.00,
        0/1/0.33, 1/1/0.29, 2/1/0.33, 3/1/0.00, 4/1/0.04,
        0/2/0.18, 1/2/0.14, 2/2/0.53, 3/2/0.09, 4/2/0.07,
        0/3/0.05, 1/3/0.08, 2/3/0.38, 3/3/0.38, 4/3/0.10,
        0/4/0.00, 1/4/0.00, 2/4/0.09, 3/4/0.16, 4/4/0.75
    }
    \node[rotate=90, font=\scriptsize\sffamily] at (-1.2, -2) {True NHI};

    \confusionmatrix{6}{0}{\FTN}{
        0/0/0.55, 1/0/0.42, 2/0/0.03, 3/0/0.00, 4/0/0.00,
        0/1/0.07, 1/1/0.75, 2/1/0.17, 3/1/0.00, 4/1/0.00,
        0/2/0.02, 1/2/0.27, 2/2/0.53, 3/2/0.16, 4/2/0.02,
        0/3/0.00, 1/3/0.00, 2/3/0.14, 3/3/0.75, 4/3/0.11,
        0/4/0.00, 1/4/0.00, 2/4/0.09, 3/4/0.14, 4/4/0.77
    }

    \confusionmatrix{12}{0}{\KNL}{
        0/0/0.91, 1/0/0.09, 2/0/0.00, 3/0/0.00, 4/0/0.00,
        0/1/0.00, 1/1/0.18, 2/1/0.36, 3/1/0.45, 4/1/0.00,
        0/2/0.00, 1/2/0.00, 2/2/0.39, 3/2/0.50, 4/2/0.11,
        0/3/0.03, 1/3/0.00, 2/3/0.03, 3/3/0.72, 4/3/0.21,
        0/4/0.00, 1/4/0.00, 2/4/0.00, 3/4/0.11, 4/4/0.89
    }

    \confusionmatrix{3}{-6.5}{Overall}{
        0/0/0.82, 1/0/0.15, 2/0/0.03, 3/0/0.00, 4/0/0.00,
        0/1/0.15, 1/1/0.52, 2/1/0.25, 3/1/0.07, 4/1/0.01,
        0/2/0.08, 1/2/0.16, 2/2/0.50, 3/2/0.20, 4/2/0.06,
        0/3/0.02, 1/3/0.02, 2/3/0.19, 3/3/0.63, 4/3/0.13,
        0/4/0.00, 1/4/0.00, 2/4/0.08, 3/4/0.14, 4/4/0.78
    }
    \begin{scope}[shift={(3,-6.5)}]
            \node[rotate=90, font=\scriptsize\sffamily] at (-1.2, -2) {True NHI};
        \node[font=\scriptsize\sffamily] at (2, -5.2) {Predicted NHI};
    \end{scope}

    \confusionmatrix{9}{-6.5}{PathAI}{
        0/0/0.80, 1/0/0.15, 2/0/0.05, 3/0/0.00, 4/0/0.00,
        0/1/0.33, 1/1/0.50, 2/1/0.17, 3/1/0.00, 4/1/0.00,
        0/2/0.02, 1/2/0.08, 2/2/0.79, 3/2/0.08, 4/2/0.03,
        0/3/0.00, 1/3/0.01, 2/3/0.24, 3/3/0.69, 4/3/0.06,
        0/4/0.00, 1/4/0.00, 2/4/0.00, 3/4/0.29, 4/4/0.71
    }
    \begin{scope}[shift={(9,-6.5)}]
        \node[font=\scriptsize\sffamily] at (2, -5.2) {Predicted NHI};
    \end{scope}

    \begin{scope}[shift={(14.5, -6.5)}]
        \shade[top color=maxblue, bottom color=minblue] (0, 0.5) rectangle (0.5, -4.5);
        
        \draw (0, 0.5) rectangle (0.5, -4.5);
        
        \foreach \val/\y in {0.0/-4.5, 0.2/-3.5, 0.4/-2.5, 0.6/-1.5, 0.8/-0.5, 1.0/0.5} {
            \draw (0.5, \y) -- (0.7, \y);
            \node[anchor=west, font=\scriptsize\sffamily] at (0.8, \y) {\val};
        }
    \end{scope}

\end{tikzpicture}
    \caption{Confusion matrices for five-grade NHI prediction: \IKEM, \FTN, and \KNL (top row), whole final test set across all centers (Overall), and PathAI reference results. The colorbar indicates normalized frequencies.}
    \label{fig:cm_by_center}
    \vspace{-3mm}
\end{figure}

In \cref{fig:cm_by_center} we report the full five-grade NHI performance on the final test sets for individual centers (\IKEM, \FTN, \KNL), on the whole final test set (Overall), and the results of PathAI, a prior multicenter AI system for UC histology.
PathAI results are taken from~\cite{Najdawi2023}.
In the PathAI work, agreement with expert grading was high (weighted Cohen's $\kappa = 0.91$; Spearman correlation $\rho = 0.89$, $P < .001$).
Our model achieves weighted $\kappa = 0.85$ and Spearman $\rho = 0.85$ ($P < .001$).
Note that performance across NHI classes varies among hospitals, which is likely driven in part by inter-observer variability in routine grading.

We conclude that our performance is comparable to strong reference systems while being trained on only case-/slide-level labels.
This weakly supervised setting substantially reduces the annotation burden on expert pathologists relative to approaches that rely on dense region- or cell-level supervision, and better matches the constraints of real-world multicenter data.

\begin{figure}[t]
\centering
\includegraphics[width=0.49\linewidth]{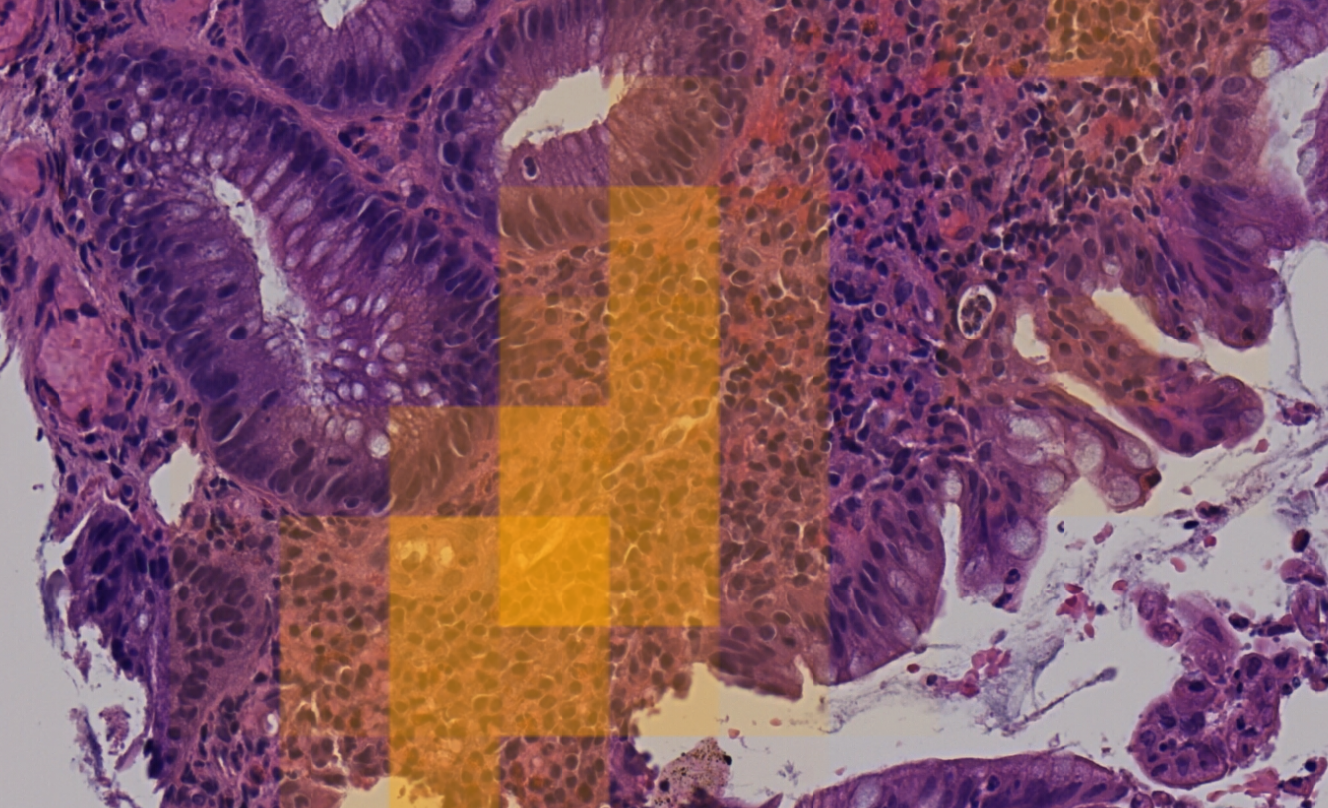}
\hfill
\includegraphics[width=0.49\linewidth]{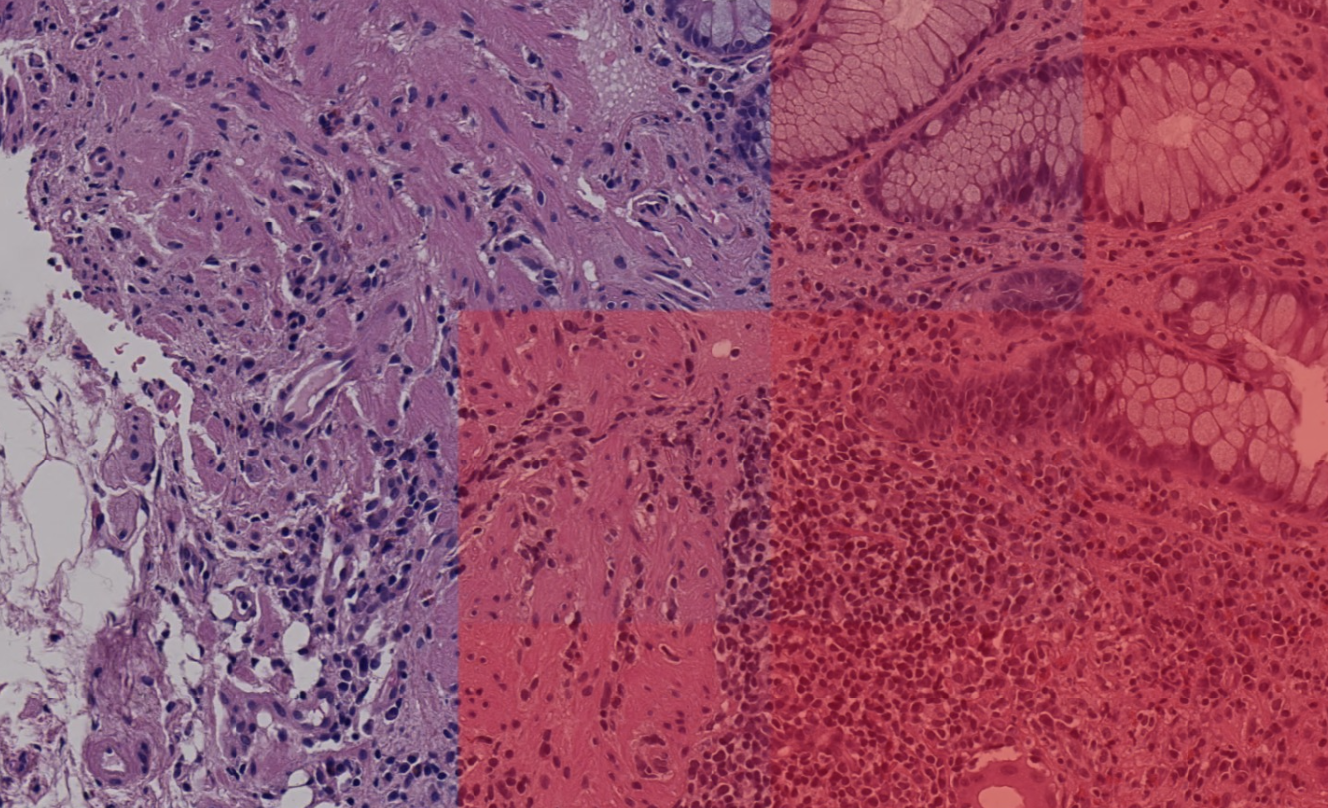}
\caption{Qualitative interpretability for a representative region: attention-weight overlay (left) and instance-first classification overlay (right). Overlays visualized in~xOpat~\cite{https://doi.org/10.1111/cgf.14812}.}
\label{fig:attn_vis}
\vspace{-3mm}
\end{figure}

\paragraph{Interpretability via Attention Visualization.}
Beyond aggregate performance, interpretability is crucial for clinical deployment.
Attention-based MIL provides an intuitive mechanism to localize evidence by highlighting tiles that contribute most to the slide-level decision.
Our instance-first formulation keeps the global decision differentiable and attention-weighted while enabling direct tile-level scoring via $c_{\psi}(h_i)$ for visualization.
Such tile-level scores improve the interpretability of the slide-level decision.
Figure~\ref{fig:attn_vis} illustrates two complementary views for the same tissue region:
an overlay of attention weights from the MIL aggregator and an overlay derived from the instance-first classification head.
Such visualizations help pathologists inspect model behavior, identify failure modes, and accelerate validation of model outputs.

\paragraph{Expert Pathologist Feedback.}
From a pathologist’s perspective, the model’s behavior appears clinically feasible.
It shows a clear ability to distinguish inactive (NHI 0) from highly active (NHI 4) cases.
Most misclassifications occur between adjacent grades, which mirrors known areas of inter-observer variability in routine practice.
Attention overlays frequently highlight regions with neutrophilic cryptitis, crypt abscesses, and surface epithelial injury, supporting biological interpretability of the predictions.
Importantly, the weakly supervised approach reflects real-world reporting conditions, where slide-level labels are available, but detailed region-level annotations are not.
Such a system could serve as a decision-support tool, standardizing grading across centers and reducing interobserver variability.

\subsection{Ablation Studies}
\paragraph {Comparison of Foundation Model Encoders.}
\label{sec:results_fm}
In our MIL pipeline, we tested three frozen foundation model encoders, UNI2-h~\cite{Chen2024}, Prov-GigaPath~\cite{Xu2024} (PGP), and Virchow2~\cite{zimmermann2024virchow2scalingselfsupervisedmixed}.
All our models were trained and evaluated using the tiling setting (T1).
In addition, we evaluated Virchow2 with an alternative tiling resolution (T2).
We report this setting only for Virchow2, as it is the only evaluated foundation model that supports multi-resolution inputs.

Table~\ref{tab:fm_all} summarizes classification performance for the three tasks (neutrophils, Nancy-low, Nancy-high), reporting accuracy, precision, recall, and specificity. 
For the multi-class tasks (Nancy-low/high), metrics are macro-averaged across all in-distribution classes (excluding \Lo{} and \Hi).

\begin{table}[tb]
    \caption{%
        Combined comparison of foundation model encoders across the three tasks used in our pipeline: neutrophil activity (binary; NHI 0--1 vs.\ 2--4), Nancy-low (multi-class), and Nancy-high (multi-class).
        Accuracy (Acc.), precision (P), recall (R), and specificity (S) are reported on the preliminary test set.
        For the multi-class tasks, P/R/S are computed using macro averaging over the in-distribution NHI grades (excluding the \Lo{} and \Hi{} classes).
    }
    \label{tab:fm_all}
    \centering
    \fontsize{8}{9.5}\selectfont
    \setlength{\tabcolsep}{4pt}
    \begin{tabular}{l|cccc|cccc|cccc}
        \toprule
        \multicolumn{1}{l}{\multirow{2}{*}[-2pt]{\textbf{Encoder}}} &
        \multicolumn{4}{c}{\textbf{Neutrophils}} &
        \multicolumn{4}{c}{\textbf{Nancy Low}} &
        \multicolumn{4}{c}{\textbf{Nancy High}}
        \\
        \cmidrule(lr){2-5}
        \cmidrule(lr){6-9}
        \cmidrule(lr){10-13}
        \multicolumn{1}{l}{}
        & Acc. & P & R & \multicolumn{1}{c}{S} 
        & Acc. & P & R & \multicolumn{1}{c}{S} 
        & Acc. & P & R & \multicolumn{1}{c}{S} 
        \\
        \midrule
        UNI2-h (T1)   & .904 & .916 & .863 & .937 & .719 & .686 & .719 & .913 & .642 & .625 & .642 & .914 \\
        PGP (T1)      & .909 & .916 & .873 & .937 & .720 & .687 & .720 & .921 & .638 & .628 & .635 & .908 \\
        Virchow2 (T1) & .906 & .907 & .878 & .929 & .750 & .705 & .744 & .930 & \textbf{.695} & .673 & .690 & .918 \\
        Virchow2 (T2) & \textbf{.910} & .914 & .925 & .893 & \textbf{.810} & .764 & .806 & .930 & .691 & .664 & .688 & .914 \\
        \bottomrule
    \end{tabular}
    \vspace{-3mm}
\end{table}

Overall, Virchow2 with the alternative tiling resolution (T2) achieved the best accuracy for the neutrophil and Nancy-low tasks.
For the Nancy-high task, Virchow2 with tiling (T1) was slightly better.
We nevertheless continued with Virchow2 and tiling (T2), as the accuracy difference was small while the computation time was approximately 10\texttimes shorter for tiling (T2).

\paragraph{Hierarchical Decision Baseline.}
We also tested a hierarchical rule-based (neutrophil gating) post-processing, in which the neutrophil model first determines whether a slide is Nancy-low or Nancy-high and then applies the corresponding specialist model.
In this setup, the specialist models can be trained without explicit \Lo/\Hi group classes.
This approach underperformed our proposed ensemble (weighted $\kappa = 0.87$ and Spearman $\rho = 0.87$ ($P < .001$) vs weighted $\kappa = 0.89$ and Spearman $\rho = 0.88$ ($P < .001$)), indicating that agreement across task-specific views yields more reliable five-grade predictions.

\paragraph{External Neutrophil Detector.}
We also evaluated a variant in which the neutrophil component was replaced by an external pretrained neutrophil detector~\cite{OHARA2025846}.
The detector operates at the tile level (T0) and outputs neutrophil detections with confidence scores. 
We aggregate detections across all tiles of a slide to obtain slide-level evidence.
However, this detector was outperformed by our neutrophil model by a large margin (precision: 0.91 vs 0.74; recall: 0.91 vs 0.73; F1-score: 0.91 vs 0.73; Spearman $\rho$: 0.82 vs 0.47 ($P < .001$)).

\vspace{-1mm}
\section{Conclusion}
\vspace{-1mm}
We addressed weakly supervised histologic activity assessment in UC from multicenter WSIs.
We proposed a MIL pipeline that combines tissue filtering and quality control with modern pathology foundation model encoders and task-specific MIL aggregation for clinically relevant endpoints, including neutrophilic activity and derived NHI-grade groupings.

Across centers with heterogeneous staining protocols, our approach achieved strong performance and benefited from leveraging transferable representations, with Virchow2 providing the most consistent gains.
Moreover, qualitative attention-based visualizations provide an interpretable view of model evidence and can support expert validation.

We further showed that our neutrophil detector outperforms existing available solution and that ensembling strategy can improve full five-grade NHI prediction compared to a hierarchical gating alternative, supporting more robust decision making around low/high activity boundaries.

\end{document}